# Modeling Electrical Daily Demand in Presence of PHEVs in Smart Grids with Supervised Learning

Marco Pellegrini, *Member, IEEE*

LIF S.r.l.
Via di Porto 159 – 50018 Scandicci (FI), Italy
marcopellegrini75@yahoo.it

Farshad Rassaei, *Student member, IEEE*

Department of Electrical and Computer Engineering
National University of Singapore, Singapore
farshad@u.nus.edu

*Abstract*—Replacing a portion of current light duty vehicles (LDV) with plug-in hybrid electric vehicles (PHEVs) offers the possibility to reduce the dependence on petroleum fuels together with environmental and economic benefits. The charging activity of PHEVs will certainly introduce new load to the power grid. In the framework of the development of a smarter grid, the primary focus of the present study is to propose a model for the electrical daily demand in presence of PHEVs charging. Expected PHEV demand is modeled by the PHEV charging time and the starting time of charge according to real world data. A normal distribution for starting time of charge is assumed. Several distributions for charging time are considered: uniform distribution, Gaussian with positive support, Rician distribution and a non-uniform distribution coming from driving patterns in real-world data. We generate daily demand profiles by using real-world residential profiles throughout 2014 in the presence of different expected PHEV demand models. Support vector machines (SVMs), a set of supervised machine learning models, are employed in order to find the best model to fit the data. SVMs with radial basis function (RBF) and polynomial kernels were tested. Model performances are evaluated by means of mean squared error (MSE) and mean absolute percentage error (MAPE). Best results are obtained with RBF kernel: maximum (worst) values for MSE and MAPE were about $2.89 \cdot 10^{-8}$ and 0.023, respectively.

## I. Introduction

The daily residential power demand profile has a high peak-to-average ratio (PAR) which significantly decreases the power grids' efficiency and causes huge costs for fortifying the power grid's infrastructure, i.e., increasing the power generation capacity, transmission lines, and distribution sector of the grid. This enormous extra investment is just to serve the power demand of the users throughout few peak demand periods. Hence, addressing this problem has motivated considerable research on techniques that can employ the current power grid more efficiently so that more consumers can be served without developing new costly infrastructure [1, 2].

When the penetration of plug-in hybrid electric vehicles (PHEVs) becomes common for residential users, we can assume that a new electricity consuming appliance is added to the houses and demands power from the grid. However, PHEVs add quite significant load to the current power grid and in particular the low voltage (LV) electricity distribution sector. Thus, the expected power demand coming from PHEVs should be investigated for proactive grid fortification. On the other hand, it is promising that unlike normal household appliances, PHEVs introduce some power demand elasticity. This elasticity comes from the facts that they do not need *on-demand* power provision and the commuting patterns differ from one household to another. Therefore, PHEVs demand can be managed to reduce their adverse impacts on the grid and LV sector especially during the peak hours [3].

Demand Response (DR) and demand side energy management (DSM) in residential sector are considered to play a key role in the smart grid framework. In the recent past many studies have been conducted on DR and DSM and reviewed in [4]. Machine learning and optimization techniques have been proposed to build a DR and home DSM system [5]. Due to the expected large quantity of PHEVs that will be integrated into the power grid in the near future and their complex charging behavior, the impact of substantial PHEVs charging on the power grid needs to be investigated [6]. Due to their ability to deal with nonlinearities of the input data, artificial neural networks (ANNs) and fuzzy logic (FL) models are commonly used techniques for modeling and forecasting load demand [7]. The main contribution of this paper is using Support vector machines (SVMs), a set of supervised machine learning algorithms, in order to model the uncoordinated charging demand of PHEVs in a residential context. The use of such algorithms has only recently been proposed for short-term load forecasting [8] and their performances are quite promising.

Real-world electrical daily demand data measured during whole year 2014 are used. The expected additive power demand from PHEVs is modeled by assigning certain distributions to the required information such as the arrival times at the households and the required charging times for respective PHEVs. A normal distribution for arrival times is assumed. Several distributions for charging time that could represent real-world vehicles' usage are considered: uniform distribution, Gaussian with positive support, Rician distribution and a non-uniform distribution which shows vehicles' mileage in real-world more accurately. The primary focus of this work is providing a unique model which is able to deal with several PHEVs charging scenarios including when no PHEV requires charging. A supervised machine learning algorithm is employed in order to find the best model to fit the data. In particular, we aim to find a single set of model parameters able to simultaneously cover all charging and weather scenarios.

## II. REGRESSION VIA SUPERVISED LEARNING

Supervised learning is the machine learning task of inferring a function from labeled training examples [9]. In machine learning, each example is a pair consisting of an input vector and a desired (target) output value (also called the supervisory vector). Support Vector Machines (SVMs) are a set of very effective supervised learning methods based on statistical learning theory originally developed by Vapnik [9]. The basic idea is to map the original input data using a nonlinear kernel function into a high dimensional feature space and determine an optimal separating hyperplane. SVMs are used for classification (SVC), regression (SVR) and outliers' detection [10].

In a classification problem, the aim is to find an optimal hyperplane that separates sample data into two classes. In a regression problem the normal to the hyperplane defines a function for which the target and the estimated values are as close as possible. In this study we used SVR in order to estimate a function based on a given training data set. Considering a set of data points $D = \{(x_1, t_1), …, (x_N, t_N)\}$ where $x_i \in R^n$ represents the input vector, $t_i \in R$ is the corresponding target and $N$ is the size of the data set, the general form of $\nu$-SVR [11] estimating function is:

$$f(x) = w^T \phi(x) + b \quad (1)$$

where $\phi(x)$ is the nonlinear map to the feature space and coefficients $w$ and $b$ are obtained by solving the following minimization problem:

$$\min \; \tfrac{1}{2} \|w\|^2 + C \left( \nu \varepsilon + 1/N \sum_N (\xi_i + \xi_i^*) \right) \quad (2)$$

subject to

$$(w^T \phi(x_i) + b) - t_i \leq \varepsilon + \xi_i,$$

$$t_i - (w^T \phi(x_i) + b) \leq \varepsilon + \xi_i^*,$$

$$\xi_i, \xi_i^* \geq 0, \; i = 1, …, N, \; C > 0, \; \varepsilon \geq 0$$

where $C$ is the regularization parameter, $\xi_i$ and $\xi_i^*$ are slack variables, $0 \leq \nu \leq 1$, and the $\varepsilon$-insensitive loss function means that no loss is assumed if $f(x)$ is in the $[t \pm \varepsilon]$ range. $\nu$ parameter is used to control the number of *support vectors*. It represents an upper bound on the fraction of training errors and a lower bound of the fraction of support vectors.

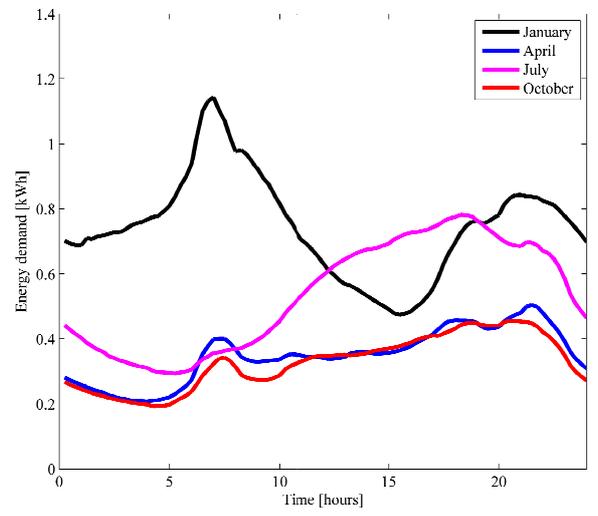

Fig. 1. Averaged electrical load measured at a residential household during four months in 2014.

Nonlinear $\nu$-SVR has also a dual formulation given by [12]:

$$\min \; \tfrac{1}{2} (\alpha - \alpha^*)^T Q (\alpha - \alpha^*) + t^T (\alpha - \alpha^*) \quad (3)$$

subject to

$$e^T(\alpha - \alpha^*) = 0,$$

$$e^T(\alpha + \alpha^*) \leq C\nu,$$

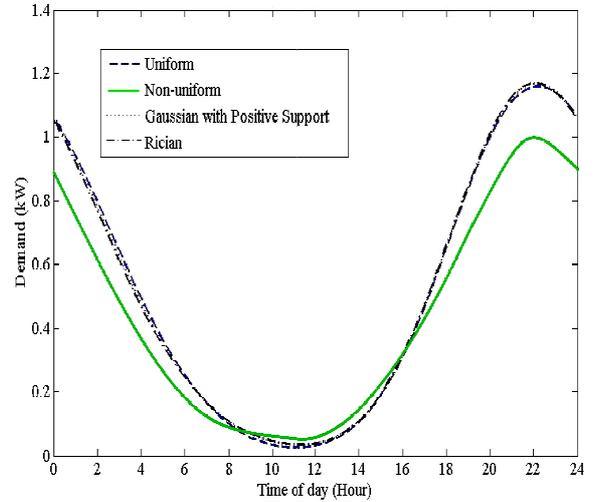

Fig. 2: Expected daily power demand of a PHEV with uniform, non-uniform, Gaussian with positive support and Rician distributions for the required charging time [13].

$$0 \leq \alpha_i, \alpha_i^* \leq C/N, \; i = 1, …, N$$

where $Q(x_i, x_j) = \phi(x_i)^T \phi(x_j)$ is the kernel, $e$ is the vector with all components equal to 1 and $\alpha_i$ and $\alpha_i^*$ are the *Lagrange* multipliers. There are number of kernels that can be used in

SVM models. These include linear, polynomial, Gaussian radial basis function (RBF) and sigmoid. RBF kernel is defined as:

$$Q(x_i, x_j) = exp(-\gamma \|x_i - x_j\|^2), \quad \gamma > 0 \quad (4)$$

where $\gamma$ represents the inverse of the radius of influence of samples selected by the model as support vectors [14]. For the present study we tested RBF and polynomial kernels, the most popular choices of kernel types used in SVMs [15].

### III. EXPECTED PHEV DEMAND MODEL

In this section the input to the SVMs model is presented. Electrical daily demand considered for this work consists of real measured data from the huge Electrical Reliability Council of Texas (ERCOT) database [16], in which power loads from clusters of households were recorded for nearly 200 different locations.

In the ERCOT database, 15-minute kWh values related to three different profile types (residential, business and non-metered) and eight weather zones are available. Besides dependence on profile type and weather zone, power demand changes depending on the time scale considered: the hour within a given day, business or non-working day, month and season [8]. ERCOT data measured at a cluster of residential households located in north Texas during four months in 2014 (January, April, July and October) were considered in order to cover all weather scenarios [17].

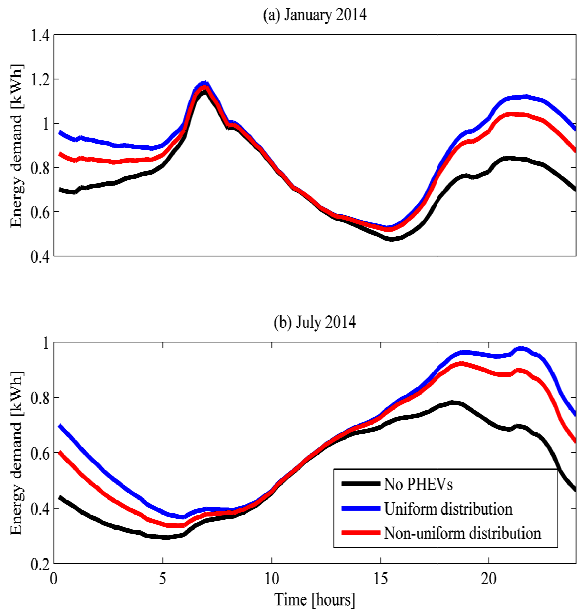

Fig. 3. Total daily demand $D_T$ evaluated by taking averaged electrical load as in Fig. 1 for (a) January and (b) July 2014 and assuming three different PHEVs' demand scenarios.

As an example, in Fig. 1 is shown the electrical load averaged over all 15-minute intervals for the above-mentioned months. It is evident from Fig. 1 the dependence of electrical demand on the season and the hour within a day. As in [18], we modeled a case in which a very low percentage of the light-duty fleet in ERCOT is PHEVs.

Without loss of generality we then assumed that the total daily demand $D_T$ in a smart grid under a high penetration of PHEVs was the sum of real-world data $R_d$ and expected PHEVs demand $E_d$:

$$D_T = R_d + E_d \quad (5)$$

where $E_d = E\{C_d(t)\}$ and $C_d(t)$ is the PHEVs' charging demand.

In order to obtain the expected power demand from PHEVs, commuting data collection by surveying can be carried out. However, in [12], it has been proven that by assigning certain distributions to the required information such as the arrival times at the households and the required charging times for respective PHEVs can provide a closed-form expression to show the uncoordinated charging demand of a PHEV mathematically. In fact, PHEVs' power demand depends on their mileage as well as their respective arrival time. Hence, we used the distributions presented in [13] and [19]. A Gaussian distribution in modulo 24 hours with $\mu = 19$ and $\sigma^2 = 10$ has been assigned to the arrival time $T_a$. A uniform distribution over the interval [1,11] and a non-uniform distribution coming from driving patterns based on National Household Travel Survey (NHTS) data [20], a Gaussian with positive support and a Rician distribution has been considered for the required charging time $T_c$. We set the parameters of the distributions such that they have the same mean and variance. The authors showed that these distributions match the practical commuting data.

The curves in Fig. 2 show the expected daily uncoordinated demand of a PHEV when the charging demand $C_d(t)$ is defined as follows [13]:

$$C_d(t) = \begin{cases} p, & T_a \leq t < T_a + T_c \\ 0, & otherwise \end{cases} \quad (6)$$

where $p$ is the outlet power delivery assumed to be constant. According to Fig. 2, we can observe that the expected daily demand from Rician and Gaussian with positive support distributions tends to the same curve as the one obtained from uniform distribution. We should notice that this happens when the first and second order statistics of these two distributions are adjusted to be the same. Thus, we only considered uniform and non-uniform distribution for $T_c$ in the present work. We then built our dataset by generating quarter-hourly values of $D_T$ for each day in considered months and for three PHEVs' demand scenarios: no PHEVs, uniform and non-uniform distribution for $T_c$.

Effectiveness of load demand fit via SVMs has been verified by minimizing the mean squared error (MSE) between the model $f(x_i)$ and the target datum $t_i$:

$$MSE = \frac{1}{N}\sum_{i=1}^{N}\left(t_i - f(x_i)\right)^2 \quad (7)$$

where $N$ is the number of 15-minute intervals in chosen data set. For all performed experiments, we quantified the prediction performance by means of MSE and mean absolute percentage error (MAPE), defined as:

$$MAPE = \frac{100}{N}\sum_{i=1}^{N}\left|\frac{(t_i - f(x_i))}{t_i}\right| \quad (8)$$

MAPE is regarded as a better error measurement than MSE because it does not accentuate large errors [21]. Model and kernel parameters were then optimized searching for the best MSE and MAPE values across generated dataset.

## IV. Model Performances

In the present section the performances of the proposed $\nu$-SVR model are reported. For the practical use of SVR, authors in [22] showed that SVMs with normalized input data into the [0, 1] range outperform those with unscaled input data. Therefore, the SVR model was fed with normalized data, and then the model outputs were returned to their original scale. The load demand model has been optimized by minimizing both the MSE and MAPE for all four months and for all charging scenarios.

In this way we looked for a single set of model and kernel parameters that was suitable for the whole year, regardless of the PHEVs charging demand. $\nu$-SVR model with RBF kernel outperformed polynomial kernels. This was not a surprising result, since it is known [23] that when the dynamics of the signal under investigation are nonlinear, SVR with RBF returns more satisfactory results than other kernels such as linear or polynomial kernel.

Table I shows the results obtained by using $\nu$-SVR model with ERCOT data measured during whole year 2014 in presence of three different expected PHEVs charging demand scenarios: no PHEV charging demand, uniform and non-uniform distribution for required charging time. A value of $p = 2$ kW was used for the outlet power delivery.

TABLE I. MODEL PERFORMANCES OF PROPOSED SVR METHOD

| PHEVs charging time distribution | Month | MSE | MAPE |
|---|---|---|---|
| No demand | January | 2.887 $10^{-8}$ | 0.0195 |
| | April | 0.288 $10^{-8}$ | 0.0127 |
| | July | 1.29 $10^{-8}$ | 0.0187 |
| | October | 0.193 $10^{-8}$ | 0.0107 |
| Uniform | January | 2.177 $10^{-8}$ | 0.0143 |
| | April | 1.016 $10^{-8}$ | 0.0174 |
| | July | 2.653 $10^{-8}$ | 0.0229 |
| | October | 0.721 $10^{-8}$ | 0.0153 |
| Non-uniform | January | 2.406 $10^{-8}$ | 0.0155 |
| | April | 1.111 $10^{-8}$ | 0.0213 |
| | July | 2.082 $10^{-8}$ | 0.0202 |
| | October | 0.56 $10^{-8}$ | 0.0169 |

The following settings for model and kernel parameters have been found to be optimal for the SVM algorithm: $C$=1000, $\nu$=0.5 and $\gamma$=10. Such values have been obtained performing a coarse/fine grid search in the parameters space [8]. The proposed method required a processing time of about 13 minutes on a commercial PC platform (2 GB RAM, 3.4 GHz CPU). Results in Table I represent the average MSE and MAPE values over the considered period given a PHEVs charging scenario.

We can summarize model performances by taking the maximum (worst) values of MSE and MAPE over all charging and weather scenarios and obtaining 2.887 $10^{-8}$ and 0.0229, respectively. Maximum MSE and MAPE values are observed during additional electric demand for heaters on winter (with no PHEVs charging demand) and for air conditioners on summer (with PHEVs charging time uniformly distributed).

It can also be seen from Fig. 3 where $D_T$, evaluated with electrical load as in Fig. 1 and assuming three different PHEVs' demand scenarios, is reported for January and July 2014. In [8], the modeling approach proposed for this work was used for one-day ahead demand forecasting based on $n_d$ days as training data; in that case main objective was to predict the load demand by minimizing $n_d$ and PHEVs penetration was not considered: MAPE values ranged from 0.81 to 1.03 depending on the season. Further work is then needed to employ SVMs for short-term load demand forecasting in presence of PHEVs penetration. This stems from the fact that MAPE = 0.9%, obtained using Wavelet Transformation Error Correction-ANN, is reported in [24] and reviewed in [7] as the best result for short-term load demand forecasting. By comparing results in [24] and [8], we can observe that, although MAPE values are within the same range of forecast error, main advantages of using SVMs rather than ANNs are that the computational complexity does not depend on the dimensionality of the input and the provided solution is global and unique [25].

## V. Conclusions

In this study, a method based on supervised machine learning was proposed in the Smart Grid framework in order to model the electrical daily demand in presence of PHEVs charging. The effectiveness of the model was tested using real-world data from a cluster of residential households and evaluated by means of MSE and MAPE. The total daily demand was assumed as the sum of real demand data and expected power demand from PHEVs. The latter was modeled by assigning several probability distributions to the starting times of charging and the required charging times. Main advantages of using the proposed method rather than other artificial intelligence algorithms such as neural networks are that the computational complexity does not depend on the dimensionality of the input, model size is automatically selected and the provided solution is global and unique.